# An Axiomatic Framework for Bayesian and Belief-function Propagation

by

Prakash P. Shenoy and Glenn Shafer

*School of Business, University of Kansas, Lawrence, Kansas, 66045-2003*

## 1. Introduction

In this paper, we present axioms for local computation in hypertrees. These axioms justify the use of local computation to find marginals for a probability distribution or belief function when the probability distribution or belief function is factored on a hypertree. The axioms are abstracted from the belief-function work of Shafer, Shenoy, and Mellouli [1987], but they apply to Bayesian probabilities as well as to belief functions.

In the Bayesian case, the factorization is usually a factorization of a joint probability distribution, perhaps into marginals and conditionals. Bayesian factorizations sometimes arise from causal models, which relate each variable to a relatively small number of immediate causes; see e.g., Pearl [1986]. Bayesian factorizations can also arise from statistical models; see e.g., Darroch, Lauritzen and Speed [1980]. Belief-function factorizations generally arise from the decomposition of evidence into independent items, each involving only a few variables. We represent each item of evidence by a belief function and combine these belief functions by Dempster's rule [Shafer 1976].

We first present our general axiomatic framework and then explain how it applies to probabilities and belief functions. Before we can present the axiomatic framework, we need to review some graph-theoretic concepts. We do this in section 2. We present the framework in section 3. We apply it to probabilities in section 4 and to belief functions in section 5.

## 2. Some Concepts from Graph Theory

Most of the concepts reviewed here have been studied extensively in the graph theory literature (see Berge [1973], Golumbic [1980], and Maier [1983]). A number of the terms we use are new, however - among them, *hypertree*, *construction sequence*, *branch*, *twig*, and *Markov tree*. A *hypertree* is what other authors have called an acyclic or decomposable hypergraph. A *construction sequence* is what other authors have called a sequence with the running intersection property. A *Markov tree* is what authors in database theory have called a join tree (see Maier [1983]). We have borrowed the term *Markov tree* from probability theory, where it means a tree of variables in which separation implies probabilistic conditional independence given the separating variables. We first used the term in a non-probabilistic context in Shafer, Shenoy, and Mellouli [1987], where we justified it in terms of a concept of qualitative independence analogous to probabilistic independence.

As we shall see, hypertrees are closely related to Markov trees. The vertices of a Markov tree are always hyperedges of a hypertree, and the hyperedges of a hypertree can always be arranged in a Markov tree.

We limit our study of hypertrees to an investigation of the properties that we need for this paper. For a more thorough study of hypertrees, using only slightly different definitions, see Lauritzen, Speed, and Vijayan [1984].

**Hypergraphs and Hypertrees.** We call a non-empty set $\mathcal{H}$ of non-empty subsets of a finite set $\mathcal{X}$ a *hypergraph* on $\mathcal{X}$. We call the elements of $\mathcal{H}$ *hyperedges*. We call the elements of $\mathcal{X}$ *vertices*.

Suppose t and b are distinct hyperedges in a hypergraph $\mathcal{H}$, $t \cap b \neq \emptyset$, and b contains every vertex of t that is contained in a hyperedge of $\mathcal{H}$ other than t; if $X \in t$ and $X \in h$, where $h \in \mathcal{H}$ and

307

h≠t, then X∈ b. Then we call t a *twig* of ℋ, and we call b a *branch* for t. A twig may have more than one branch.

We call a hypergraph a *hypertree* if there is an ordering of its hyperedges, say $h_1 h_2 ... h_n$, such that $h_k$ is a twig in the hypergraph $\{h_1, h_2, ..., h_k\}$ whenever $2 \leq k \leq n$. We call any such ordering of the hyperedges a *hypertree construction sequence* for the hypertree.

In general, a hypertree will have many construction sequences. In fact, for each hyperedge of a hypertree, there is a construction sequence beginning with that hyperedge.

**Hypertree Covers of Hypergraphs.** As we will show, local computation requires two things. The joint probability distribution or belief function with which we are working must factor into functions each involving a small set of variables. And these sets of variables must form a hypertree.

If the sets of variables form instead a hypergraph that is not a hypertree, then we must enlarge it until it is a hypertree. We can talk about this enlargement in two different ways. We can say we are adding larger hyperedges, keeping the hyperedges already there. Or, alternatively, we can say we are replacing the hyperedges already there with larger hyperedges. The choice between these two ways of talking does not matter much, because the presence of superfluous twigs (hyperedges contained in other hyperedges) does not affect whether a hypergraph is a hypertree, and because the computational cost of the procedures we will be describing depends primarily on the size of the largest hyperedges, not on the number of the smaller hyperedges [Kong 1986].

We will say that a hypergraph ℋ* *covers* a hypergraph ℋ if for every h in ℋ there is an element h* of ℋ* such that h⊆h*. We will say that ℋ* is a *hypertree cover* for ℋ if ℋ* is a hypertree and it covers ℋ.

Finding a hypertree cover is never difficult. The hypertree {𝔛}, which consists of the single hyperedge 𝔛, is a hypertree cover for any hypergraph on 𝔛. Finding a hypertree cover without large hyperedges, or finding a hypertree cover whose largest hyperedge is as small as possible, may be very difficult. How to do this best is the subject of a growing literature; see e.g., Rose [1970], Bertele and Brioschi [1972], Tarjan and Yannakakis [1984], Kong [1986], Arnborg, Corneil and Proskurowski [1987], Mellouli [1987], Dechter and Pearl [1987], and Zhang [1988].

**Trees.** A *graph* is a pair $(𝒱, ℰ)$, where $𝒱$ is a non-empty set and $ℰ$ is a set of two-element subsets of $𝒱$. We call the elements of $𝒱$ *vertices*, and we call the elements of $ℰ$ *edges*.

Suppose $(𝒱, ℰ)$ is a graph. If $\{v, v'\}$ is an element of $ℰ$, then we say that v and v' are *neighbors*, or that they are *connected by an edge*. If $v_1 v_2 ... v_n$ is a sequence of distinct vertices, where n>1, and $\{v_k, v_{k+1}\} \in ℰ$ for k=1,2,...,n-1, then we call $v_1 v_2 ... v_n$ a *path*. If v and v' are distinct elements of $𝒱$, and there is a path $v_1 v_2 ... v_n$ such that $v = v_1$ and $v' = v_n$, then we say that v and v' are *connected by the path*. If every two distinct elements of $𝒱$ are connected by at least one path, then we say that $(𝒱, ℰ)$ is *connected*. If $v_1 v_2 ... v_n$ is a path, n>2, and $\{v_n, v_1\} \in ℰ$, then we call $v_1 v_2 ... v_n$ a *cycle*.

We call a vertex of a graph a *leaf* if it is contained in only one edge.

A *tree* is a graph that is connected and that has no cycles.

**Markov Trees.** We have just defined a tree as a pair $(𝒱, ℰ)$, where $𝒱$ is the set of vertices, and $ℰ$ is the set of edges. In the case of a Markov tree, the vertices are themselves non-empty sets. In other words, the set $𝒱$ is a hypergraph. In fact, it turns out to be a hypertree.

Here is our full definition. We call a tree $(ℋ, ℰ)$ a *Markov tree* if the following conditions are satisfied:

  (i) ℋ is a hypergraph.
  (ii) If $\{h, h'\} \in ℰ$, then $h \cap h' \neq \emptyset$.

(iii) If h and h' are distinct vertices, and X is in both h and h', then X is in every vertex on the path from h to h'.

Our definition does not state that $\mathcal{H}$ is a hypertree, but it implies that it is:

*Proposition 2.1.* Suppose $(\mathcal{H},\mathcal{E})$ is a Markov tree. Then $\mathcal{H}$ is a hypertree, and any leaf in $(\mathcal{H},\mathcal{E})$ is a twig in $\mathcal{H}$.

The key point here is the fact that a leaf in the Markov tree is a twig in the hypertree. This means that as we delete leaves from a Markov tree (a visually transparent operation), we are deleting twigs from the hypertree.

## 3. An Axiomatic Framework for Local Computation

In this section, we describe a general axiomatic framework that captures the essential features that makes exact local computation possible.

We define two primitive operators, marginalization and combination. These operators operate on objects called valuations. We state axioms for these operators, and we derive the possibility of local computation for the axioms.

Next, we describe a propagation scheme for computing marginals of a valuation when we have a factorization of the valuation on a hypertree.

### 3.1. The Axiomatic Framework

The framework has objects called valuations and two primitive operators, marginalization, and combination.

**Valuations.** Let $\mathcal{X}$ be a finite set. For each $h \subseteq \mathcal{X}$, there is a set $\mathcal{V}_h$. The elements of $\mathcal{V}_h$ are called *valuations on h*. Let $\mathcal{V}$ denote $\cup \{\mathcal{V}_h | h \subseteq \mathcal{X}\}$, the set of all valuations.

**Marginalization.** For each $h \subseteq \mathcal{X}$, there is a mapping $\downarrow h : \cup \{\mathcal{V}_g | g \supseteq h\} \to \mathcal{V}_h$, called *marginalization to h*, such that if G is a valuation on g and $h \subseteq g$, then $G^{\downarrow h}$ is a valuation on h representing the *marginal of G on h*.

**Combination.** There is a mapping $\otimes : \mathcal{V} \times \mathcal{V} \to \mathcal{V}$, called *combination*, such that if G and H are valuations on g and h respectively, then $G \otimes H$ is a valuation on $g \cup h$ representing the *combination of G and H*.

We will assume that these two mappings satisfy four axioms.

**Axiom A0** (*Identity*): Suppose G is a valuation on g. Then $G^{\downarrow g} = G$.

**Axiom A1** (*Consonance of marginalization*): Suppose G is a valuation on g, and suppose $h_1 \subseteq h_2 \subseteq g$. Then $G^{\downarrow h_1} = (G^{\downarrow h_2})^{\downarrow h_1}$.

**Axiom A2** (*Commutativity and associativity of combination*): Suppose G, H, K are valuations on g, h, and k respectively. Then $G \otimes H = H \otimes G$ and $G \otimes (H \otimes K) = (G \otimes H) \otimes K$.

**Axiom A3** (*Distributivity of marginalization over combination*): Suppose G is a valuation on g and H is a valuation on h. Then $(G \otimes H)^{\downarrow g} = G \otimes (H^{\downarrow g \cap h})$.

One implication of Axiom A2 is that when we have multiple combinations of valuations, we can write it without using parenthesis. For example, $(...((A_{h_1} \otimes A_{h_2}) \otimes A_{h_3}) \otimes ... \otimes A_{h_n})$ can be written simply as $\otimes \{A_{h_i} | i=1,...,n\}$ without indicating the order in which the combinations are carried out.

**Factorization.** Suppose A is a valuation on a finite set of variables $\mathcal{X}$, and suppose $\mathcal{H}$ is a hypergraph on $\mathcal{X}$. If A is equal to the combination of valuations on the hyperedges of h, say $A = \otimes \{A_h | h \in \mathcal{H}\}$, where $A_h$ is a valuation on h, then we say that A *factorizes on* $\mathcal{H}$.

The following proposition, which follows directly from axiom A3, is the key to local propagation on hypertrees.

*Proposition 3.1.* Suppose A is a valuation on $\mathscr{X}$, suppose A factorizes on a hypergraph $\mathscr{H}$, i.e., $A = \otimes\{A_h | h \in \mathscr{H}\}$, and suppose t is a twig in $\mathscr{H}$ with branch b. Let $\mathscr{H}'$ denote $\mathscr{H}-\{t\}$ and let $\mathscr{X}'$ denote $\cup\mathscr{H}' = \mathscr{X}-(t-b)$. Then the marginal of A to $\mathscr{X}'$ factorizes on $\mathscr{H}'$ as follows:

$$A^{\downarrow \mathscr{X}'} = \otimes\{A_h \mid h \in \mathscr{H}-\{t,b\}\} \otimes (A_b \otimes A_t^{\downarrow t \cap b}) \qquad (3.1)$$

Proposition 3.1 tells us that when a valuation factorizes on a hypergraph, marginalization of the valuation by removal of a twig can be done *locally* in the sense that only the valuations on the twig and its branch are involved.

Proposition 3.1 is especially interesting in the case of hypertrees, because repeated application of (3.1) allows us to obtain A's marginal on any particular hyperedge of $\mathscr{H}$. If we want the marginal on a hyperedge $h_1$, we choose a construction sequence beginning with $h_1$, say $h_1 h_2 ... h_n$. Suppose $\mathscr{X}_k$ denotes $h_1 \cup ... \cup h_k$ and $\mathscr{H}_k$ denotes $\{h_1, h_2, ..., h_k\}$ for $k=1,...,n-1$. Then axiom A1 tells us that $A^{\downarrow h_1} = (...(A^{\downarrow \mathscr{X}_{n-1}})^{\downarrow \mathscr{X}_{n-2}}...)^{\downarrow \mathscr{X}_1}$. We use (3.1) to delete the twig $h_n$, so that we have a factorization of $A^{\downarrow \mathscr{X}_{n-1}}$ on the hypertree $\mathscr{H}_{n-1}$. Then we use (3.1) again to delete the twig $h_{n-1}$ so that we have a factorization of $A^{\downarrow \mathscr{X}_{n-2}}$ on the hypertree $\mathscr{H}_{n-2}$. And so on, until we have deleted all the hyperedges except $h_1$, so that we have a factorization of $A^{\downarrow h_1}$ on the hypertree $\{h_1\}$, i.e., we have the marginal $A^{\downarrow h_1}$. At each step, the computation is local, in the sense that it involves only a twig and its branch.

### 3.2. The Propagation Scheme

In this section, we translate our marginalization scheme from the hypertree to the Markov tree, change metaphors from deletion of leaves to message passing, and finally generalize the scheme so that we can compute marginals for all the vertices of a Markov tree.

Successively deleting hyperedges in a hypertree until only a single hyperedge remains translates to successively deleting leaves in the Markov tree until only a single vertex remains.

We now change the metaphor from deletion of leaves to message passing. Why did we talk about deleting the hyperedge $h_k$ as we marginalized $h_k$'s potential to its intersection with its branch? The point was simply to remove $h_k$ from our attention. The deletion had no computational significance, but it helped to make clear that $h_k$ and its potential were of no further use. What was of further use was the smaller hypertree that would remain were $h_k$ deleted.

When we turn from the hypertree to the Markov tree, since a tree is easier to visualize than a hypertree, we can remove a leaf from our attention without leaning so heavily on the metaphorical deletion. And a Markov tree also allows another, more useful, metaphor. We can imagine that each vertex of the tree is a processor, and we can imagine that the marginalization is a message that one processor passes to another. In terms of the message passing metaphor, to obtain the marginal for a given vertex, we need only to send messages inwards towards that vertex starting from leaves. Regarding timing, each processor waits till it has received a message from all its outward neighbors before sending a message to its inward neighbor.

Now suppose we wish to compute marginals for all vertices of the Markov tree simultaneously. Each processor now sends messages to all its neighbors. Regarding timing, a processor sends a message to a neighbor only after the processor has received a message from all its other neighbors.

We can describe this scheme for computing marginals on all vertices simultaneously in terms of a forward-chaining production system. A forward-chaining production system consists of a working memory and a rule-base, a set of rules for changing the contents of the memory. (See Brownston *et al* [1985].)

310

Let A be a valuation on $\mathfrak{X}$. Suppose that we have a factorization of A on a hypertree $\mathcal{H}$ on $\mathfrak{X}$, i.e., $A = \otimes\{A_{h_i} | h_i \in \mathcal{H}\}$. We wish to compute $A^{\downarrow h_i}$, the marginal of A for each hyperedge $h_i$ of $\mathcal{H}$. Suppose that the mappings $\downarrow h$ and $\otimes$ satisfy Axioms A0 to A3.

Let $\mathfrak{M}$ be a Markov tree for the hypertree $\mathcal{H}$. Let $\mathfrak{N}_i$ denote the set of all neighbors of vertex i in $\mathfrak{M}$. We will imagine that there is an independent processor at each vertex of the Markov tree and that these processors are connected in the same way as the vertices are in the Markov tree. In the scheme, each vertex i will transmit a valuation to each of its neighbor. The valuation transmitted by vertex i to its neighboring vertex j will be denoted by $M^{i \to j}$.

We start with a working memory that contains $A_{h_i}$ for each vertex i of $\mathfrak{M}$. The rule base has just two rules:

**Rule 1**: If $j \in \mathfrak{N}_i$ and $M^{k \to i}$ is present in working memory for all k in $\mathfrak{N}_i - \{j\}$, then compute $M^{i \to j}$ as follows:
$$M^{i \to j} = (A_{h_i} \otimes (\otimes\{M^{k \to i} | k \in \mathfrak{N}_i - \{j\}\}))^{\downarrow h_j}$$
and place it in working memory.

**Rule 2**: If $M^{k \to i}$ is present in working memory for all k in $\mathfrak{N}_i$, then compute $A^{\downarrow h_i}$ as follows:
$$A^{\downarrow h_i} = A_{h_i} \otimes (\otimes\{M^{k \to i} | k \in \mathfrak{N}_i\})$$
and print the result.

Rule 1 will initially fire just for edges $\{i,j\}$ such that i is a leaf. Rule 1 will eventually fire in both directions for each edge $\{i,j\}$ in $\mathfrak{M}$ producing $M^{i \to j}$ and $M^{j \to i}$. Rule 2 will eventually fire once for each vertex i of $\mathfrak{M}$.

## 4. Probability Propagation

In this section, we explain local computation for probability distributions. More precisely, we show how the problem of computing marginals of joint probability distributions fits the general framework described in the previous section.

For probability propagation, valuations will correspond to potentials.

**Potentials**. We use the symbol $\mathcal{W}_X$ for the set of possible values of a variable X, and we call $\mathcal{W}_X$ the *frame* for X. We will be concerned with a finite set $\mathfrak{X}$ of variables, and we will assume that all the variables in $\mathfrak{X}$ have finite frames. For each $h \subseteq \mathfrak{X}$, we let $\mathcal{W}_h$ denote the Cartesian product of $\mathcal{W}_X$ for X in h; we call $\mathcal{W}_h$ the *frame* for h. We will refer to elements of $\mathcal{W}_h$ as *configurations of h*. A *potential* on h is a real-valued function on $\mathcal{W}_h$ that has non-negative values that are not all zero. Intuitively, potentials are unnormalized probability distributions.

**Marginalization**. Marginalization is familiar in probability theory; it means reducing a function on one set of variables to a function on a smaller set of variables by summing over the variables omitted.

In order to develop a notation for the marginalization of potentials, we first need a notation for the marginalization of configurations of a set of variables to a smaller set of variables. Here marginalization simply means dropping extra coordinates; if (w,x,y,z) is a configuration of $\{W,X,Y,Z\}$, for example, then the marginalization of (w,x,y,z) to $\{W,X\}$ is simply (w,x), which is a configuration of $\{W,X\}$. If g and h are sets of variables, $g \subseteq h$, and **x** is a configuration of h, then we will let $\mathbf{x}^{\downarrow g}$ denote the marginalization of **x** to g.

Suppose g and h are sets of variables, $h \subseteq g$, and G is a potential on g. The *marginal of G to h*, denoted by $G^{\downarrow h}$, is a potential on h defined by

311

$$G^{\downarrow h}(\mathbf{x}) = \begin{cases} \Sigma\{G(\mathbf{y}) | \mathbf{y} \in \mathcal{W}_g \text{ and } \mathbf{y}^{\downarrow h} = \mathbf{x}\} & \text{if } h \neq g \\ G(\mathbf{x}) & \text{if } h = g \end{cases}$$

for all $\mathbf{x} \in \mathcal{W}_h$.

It is obvious from the above definition that marginalization operation for potentials satisfies axioms A0 and A1.

**Combination.** For potentials, combination is simply pointwise multiplication. If G is a potential on g, H is a potential on h, and there exists an $\mathbf{x} \in \mathcal{W}_{g \cup h}$ such that $G(\mathbf{x}^{\downarrow g})H(\mathbf{x}^{\downarrow h}) > 0$, then their *combination*, denoted simply by GH, is the potential on $g \cup h$ given by $(GH)(\mathbf{x}) = G(\mathbf{x}^{\downarrow g})H(\mathbf{x}^{\downarrow h})$ for all $\mathbf{x} \in \mathcal{W}_{g \cup h}$. If there exists no $\mathbf{x} \in \mathcal{W}_{g \cup h}$ such that $G(\mathbf{x}^{\downarrow g})H(\mathbf{x}^{\downarrow h}) > 0$, then the combination of G and H will be undefined.

It is clear from the definition of combination of potentials that it satisfies axiom A2. It is shown in Shafer and Shenoy [1988] that the marginalization and combination operations for potentials satisfies axiom A3. Thus all axioms are satisfied making local computation possible.

A number of authors who have studied local computation for probability, including Kelly and Barclay [1973], Cannings, Thompson and Skolnick [1978], Pearl [1986], Shenoy and Shafer [1986], and Lauritzen and Spiegelhalter [1988], have described schemes that are variations on the the basic scheme described in section 3.2. Most of these authors, however, have justified their schemes by emphasizing conditional probability. We believe this emphasis is misplaced. What is essential to local computation is a factorization. It is not essential that this factorization be interpreted, at any stage, in terms of conditional probabilities. For more regarding this point, see Shafer and Shenoy [1988].

## 5. Belief-function Propagation

In this section, we explain local computation for belief functions. More precisely, we show how the problem of computing marginals of a joint belief function fits the general framework described in the previous section.

For belief-function propagation, valuations correspond to belief functions. Before we define a belief function, we need the concept of a random non-empty subset.

**Random Non-empty Subset.** Suppose $\mathcal{W}_h$ is the frame for a set of variables h. A *random subset* $\mathcal{S}$ *of* $\mathcal{W}_h$ is defined by giving a probability measure on the set of all subsets of $\mathcal{W}_h$. In other words, we assign to the subsets of $\mathcal{W}_h$ non-negative numbers adding to one. We write $\Pr[\mathcal{S}=A]$ for the non-negative number assigned to the subset A of $\mathcal{W}_h$, and we call $\Pr[\mathcal{S}=A]$ the probability that $\mathcal{S}$ is equal to A. If $\Pr[\mathcal{S}=\varnothing] = 0$, then we say that the random subset $\mathcal{S}$ is *non-empty*.

**Belief Function.** A function Bel that assigns a degree of belief Bel(A) to every subset A of $\mathcal{W}_h$ is called a *belief function* on h if there exists a random non-empty subset $\mathcal{S}$ of $\mathcal{W}_h$ such that Bel is given by $\text{Bel}(A) = \Pr[\mathcal{S} \subseteq A]$ for every subset A of $\mathcal{W}_h$. Intuitively, the number Bel(A) is the degree to which we judge given evidence to support the proposition that the true value of variables in h is in A, or the degree to which we think it reasonable to believe this proposition on the basis of that evidence alone.

**Marginalization.** If g and h are sets of variables, $h \subseteq g$, and G is a non-empty subset of $\mathcal{W}_g$, then the *marginal* of G to h, denoted by $G^{\downarrow h}$, is a subset of $\mathcal{W}_h$ given by $G^{\downarrow h} = \{\mathbf{x}^{\downarrow h} \mid \mathbf{x} \in G\}$.

For example, If A is subset of $\mathcal{W}_{\{W,X,Y,Z\}}$, then the marginal of A to $\{X,Y\}$ consists of the elements of $\mathcal{W}_{\{X,Y\}}$ which can be obtained by marginalizing elements of A to $\mathcal{W}_{\{X,Y\}}$.

Suppose Bel is a belief function on g corresponding to random non-empty subset $\mathcal{S}_g$ of $\mathcal{W}_g$. The *marginal of Bel to h*, denoted by $Bel^{\downarrow h}$, is the belief function on h corresponding to the random non-empty subset $\mathcal{S}_g^{\downarrow h}$.

We are using standard probability notation here. The random non-empty subset $\mathcal{S}_g^{\downarrow h}$ is a "function" of the random non-empty subset $\mathcal{S}_g$ in the sense that whenever $\mathcal{S}_g = G$, $\mathcal{S}_g^{\downarrow h} = G^{\downarrow h}$. Thus $\mathcal{S}_g^{\downarrow h}$ is a well-defined random non-empty subset of $\mathcal{W}_h$.

It is easy to verify that the above definition of marginalization of belief functions satisfies axioms A0 and A1.

Before we can define the combination operation for belief functions, we need the operation of vacuous extension of subsets.

**Vacuous Extension.** By vacuous extension of a subset of a frame to a subset of a larger frame, we mean a cylinder set extension. If g and h are sets of variables, $g \subseteq h$, $g \neq h$, and G is a subset of $\mathcal{W}_g$, then the *vacuous extension* of G to $\mathcal{W}_h$ is $G \times \mathcal{W}_{h-g}$. If G is a subset of $\mathcal{W}_g$, then the vacuous extension of G to $\mathcal{W}_g$ is defined to be G. We will let $G^{\uparrow h}$ denote the vacuous extension of G to $\mathcal{W}_h$.

For example, if A is a subset of $\mathcal{W}_{\{W,X\}}$, for example, then the vacuous extension of A to $\mathcal{W}_{\{W,X,Y,Z\}}$ is $A \times \mathcal{W}_{\{Y,Z\}}$.

**Combination.** Dempster's rule of combination is a rule for combining belief functions. Consider two random non-empty subsets $\mathcal{S}_g$ and $\mathcal{S}_h$ of $\mathcal{W}_g$ and $\mathcal{W}_h$ respectively. Suppose $\mathcal{S}_g$ and $\mathcal{S}_h$ are probabilistically independent, i.e., $Pr[\mathcal{S}_g = G \text{ and } \mathcal{S}_h = H] = Pr[\mathcal{S}_g = G] Pr[\mathcal{S}_h = H]$ for all subsets G of $\mathcal{W}_g$ and H of $\mathcal{W}_h$. Suppose also that $Pr[\mathcal{S}_g^{\uparrow g \cup h} \cap \mathcal{S}_h^{\uparrow g \cup h} \neq \varnothing] > 0$. Let $\mathcal{S}$ be the random non-empty subset of $\mathcal{W}_{g \cup h}$ that has the probability distribution of $\mathcal{S}_g^{\uparrow g \cup h} \cap \mathcal{S}_h^{\uparrow g \cup h}$ conditional on $\mathcal{S}_g^{\uparrow g \cup h} \cap \mathcal{S}_h^{\uparrow g \cup h} \neq \varnothing$, i.e.,
$$Pr[\mathcal{S} = A] = Pr[\mathcal{S}_g^{\uparrow g \cup h} \cap \mathcal{S}_h^{\uparrow g \cup h} = A] / Pr[\mathcal{S}_g^{\uparrow g \cup h} \cap \mathcal{S}_h^{\uparrow g \cup h} \neq \varnothing]$$
for every non-empty subset A of $\mathcal{W}_{g \cup h}$. If $Bel_g$ and $Bel_h$ are belief functions for h and g corresponding to $\mathcal{S}_g$ and $\mathcal{S}_h$ respectively, then the *combination* of $Bel_g$ and $Bel_h$, denoted by $Bel_g \oplus Bel_h$, is the belief function for $g \cup h$ corresponding to $\mathcal{S}$.

If the bodies of evidence on which $Bel_g$ and $Bel_h$ are based are independent, then $Bel_g \oplus Bel_h$ is supposed to represent the result of pooling these two bodies of evidence.

It is shown in Shafer [1976] that Dempster's rule of combination is commutative and associative. In Shafer and Shenoy [1988], it is shown that the above definitions of marginalization and combination for belief functions satisfies axiom A3. Thus all axioms are satisfied making local computation possible.

Propagation of belief functions using local computation has been studied by Shafer and Logan [1987], Shenoy and Shafer [1986], Kong [1986], Dempster and Kong [1986], Shafer, Shenoy and Mellouli [1987], Mellouli [1987] and Shafer and Shenoy [1988]. Shafer, Shenoy and Srivastava [1988], and Zarley, Hsia and Shafer [1988] discuss implementations of these propagation schemes.

### References

Arnborg, S., Corneil, D. G. and Proskurowski, A. (1987), Complexity of finding embeddings in a k-tree, *SIAM Journal of Algebraic and Discrete Methods*, **8**, 277-284.

Berge, C. (1973), *Graphs and Hypergraphs*, translated from French by E. Minieka, North-Holland.

Bertele, U. and Brioschi, F. (1972), *Nonserial Dynamic Programming*, Academic Press.

Brownston, L. S., Farrell, R. G., Kant, E. and Martin, N. (1985), *Programming Expert Systems in OPS5: An Introduction to Rule-Based Programming*, Addison-Wesley.

Cannings, C., Thompson, E. A. and Skolnick, M. H. (1978), Probability functions on complex pedigrees, *Advances in Applied Probability*, **10**, 26-61.

Darroch, J. N., Lauritzen, S. L. and Speed, T. P. (1980), Markov fields and log-linear models for contingency tables, *Annals of Statistics*, **8**, 522-539.

Dechter, R. and Pearl, J. (1987), Tree-clustering schemes for constraint processing, Cognitive Systems Laboratory Report R-92, University of California at Los Angeles.

Dempster, A. P. and Kong, A. (1986), Uncertain evidence and artificial analysis, Research Report S-108, Department of Statistics, Harvard University.

Golumbic, M. C. (1980), *Algorithmic Graph Theory and Perfect Graphs*, Academic Press.

Kelly, C. W. III and Barclay, S. (1973), A general Bayesian model for hierarchical inference, *Organizational Behavior and Human Performance*, **10**, 388-403.

Kong, A. (1986), Multivariate belief functions and graphical models, doctoral dissertation, Department of Statistics, Harvard University.

Lauritzen, S. L., Speed, T. P. and Vijayan, K. (1984), Decomposable graphs and hypergraphs, *Journal of the Australian Mathematical Society*, series A, **36**, 12-29.

Lauritzen, S. L. and Spiegelhalter, D. J. (1988), Local computations with probabilities on graphical structures and their application to expert systems (with discussion), *Journal of the Royal Statistical Society*, series B, **50**, to appear.

Maier, D. (1983), *The Theory of Relational Databases*, Computer Science Press.

Mellouli, K. (1987), On the propagation of beliefs in networks using the Dempster-Shafer theory of evidence, doctoral dissertation, School of Business, University of Kansas.

Pearl, J. (1986), Fusion, propagation and structuring in belief networks, *Artificial Intelligence*, **29**, 241-288.

Rose, D. J. (1970), Triangulated graphs and the elimination process, *Journal of Mathematical Analysis and Applications*, **32**, 597-609.

Shafer, G. (1976), *A Mathematical Theory of Evidence*. Princeton University Press.

Shafer, G. and Logan, R. (1987), Implementing Dempster's rule for hierarchical evidence, *Artificial Intelligence*, **33**, 271-298.

Shafer, G. and Shenoy, P. P. (1988), Local computation in hypertrees, School of Business Working Paper No. 201, University of Kansas.

Shafer, G., Shenoy, P. P. and Mellouli, K. (1987), Propagating belief functions in qualitative Markov trees, *International Journal of Approximate Reasoning*, **1**(4), 349-400.

Shafer, G., Shenoy, P. P. and Srivastava, R. P. (1988), AUDITOR'S ASSISTANT: A knowledge engineering tool for audit decisions, School of Business Working Paper No. 197, University of Kansas.

Shenoy, P. P. and Shafer, G. (1986), Propagating belief functions using local computations, *IEEE Expert*, **1**(3), 43-52.

Tarjan, R. E. and Yannakakis, M. (1984), Simple linear time algorithms to test chordality of graphs, test acyclicity of hypergraphs, and selectively reduce acyclic hypergraphs, *SIAM Journal of Computing*, **13**, 566-579.

Zarley, D. K., Hsia, Y. and Shafer, G. (1988), Evidential reasoning using DELIEF, School of Business Working Paper No. 193, University of Kansas.

Zhang, L. (1988), Studies on finding hypertree covers for hypergraphs, School of Business Working Paper No. 198, University of Kansas.